\documentclass[11pt]{article}

\usepackage{acl}

\usepackage{times}
\usepackage{latexsym}

\usepackage[T1]{fontenc}

\usepackage[utf8]{inputenc}

\usepackage{microtype}

\usepackage{inconsolata}

\usepackage{graphicx}

\usepackage{booktabs}
\usepackage{multirow}
\usepackage[table]{xcolor}
\usepackage{amsmath}

\usepackage{pifont}
\newcommand{\cmark}{\ding{51}} 
\newcommand{\xmark}{\ding{55}} 

\usepackage{fvextra}
\DefineVerbatimEnvironment{PromptVerbatim}{Verbatim}{breaklines=true,breakanywhere=true,fontsize=\small,xleftmargin=0pt}

\title{AdaMem: Adaptive User-Centric Memory for Long-Horizon Dialogue Agents}

\author{
 \textbf{Shannan Yan\textsuperscript{1,2}*},
 \textbf{Jingchen Ni\textsuperscript{1}*},
 \textbf{Leqi Zheng\textsuperscript{1}*},
 \textbf{Jiajun Zhang},
\\
 \textbf{Peixi Wu\textsuperscript{2}},
 \textbf{Dacheng Yin\textsuperscript{2}},
 \textbf{Jing LYU\textsuperscript{2}},
 \textbf{Chun Yuan\textsuperscript{1}$^\dagger$},
 \textbf{Fengyun Rao\textsuperscript{2}$^\dagger$}
\\
\\
 \textsuperscript{1}Tsinghua University
 \textsuperscript{2}WeChat Vision, Tencent Inc.
\\
 \small{
 }
}

\begin{document}
\maketitle

\newcommand\blfootnote[1]{%
  \begingroup
  \renewcommand\thefootnote{}\footnote{#1}%
  \addtocounter{footnote}{-1}%
  \endgroup
}
\blfootnote{*Equal Contribution. $^\dagger$Corresponding author. This work was done when Shannan Yan was an intern at Tencent Inc.} 

\vspace{-1.5em}

\begin{abstract}
Large language model (LLM) agents increasingly rely on external memory to support long-horizon interaction, personalized assistance, and multi-step reasoning. However, existing memory systems still face three core challenges: they often rely too heavily on semantic similarity, which can miss evidence crucial for user-centric understanding; they frequently store related experiences as isolated fragments, weakening temporal and causal coherence; and they typically use static memory granularities that do not adapt well to the requirements of different questions. We propose AdaMem, an adaptive user-centric memory framework for long-horizon dialogue agents. AdaMem organizes dialogue history into working, episodic, persona, and graph memories, enabling the system to preserve recent context, structured long-term experiences, stable user traits, and relation-aware connections within a unified framework. At inference time, AdaMem first resolves the target participant, then builds a question-conditioned retrieval route that combines semantic retrieval with relation-aware graph expansion only when needed, and finally produces the answer through a role-specialized pipeline for evidence synthesis and response generation. We evaluate AdaMem on the LoCoMo and PERSONAMEM benchmarks for long-horizon reasoning and user modeling. Experimental results show that AdaMem achieves state-of-the-art performance on both benchmarks. The code will be released upon acceptance.

\end{abstract}

\section{Introduction}

Recent advances in large language model (LLM)-based agents have enabled increasingly capable systems for open-ended dialogue, multi-step reasoning, and interactive assistance \citep{schlegel2025large,schmidgall2025agent}. Yet these settings are inherently long-horizon: an agent must continually accumulate information across many turns, preserve salient details as user goals evolve, and recover the right evidence when it becomes relevant again. This makes memory a central requirement rather than a peripheral add-on. A useful memory system should not only store past interactions, but also organize them in a form that remains queryable, coherent, and robust as the conversation grows longer and more diverse. Otherwise, memory can become redundant, fragmented, or misaligned with the needs of downstream reasoning, leading to inconsistent behavior and poorly grounded responses.

\begin{figure}
\vspace{-1.5em}
\centering
\includegraphics[width=0.48\textwidth]{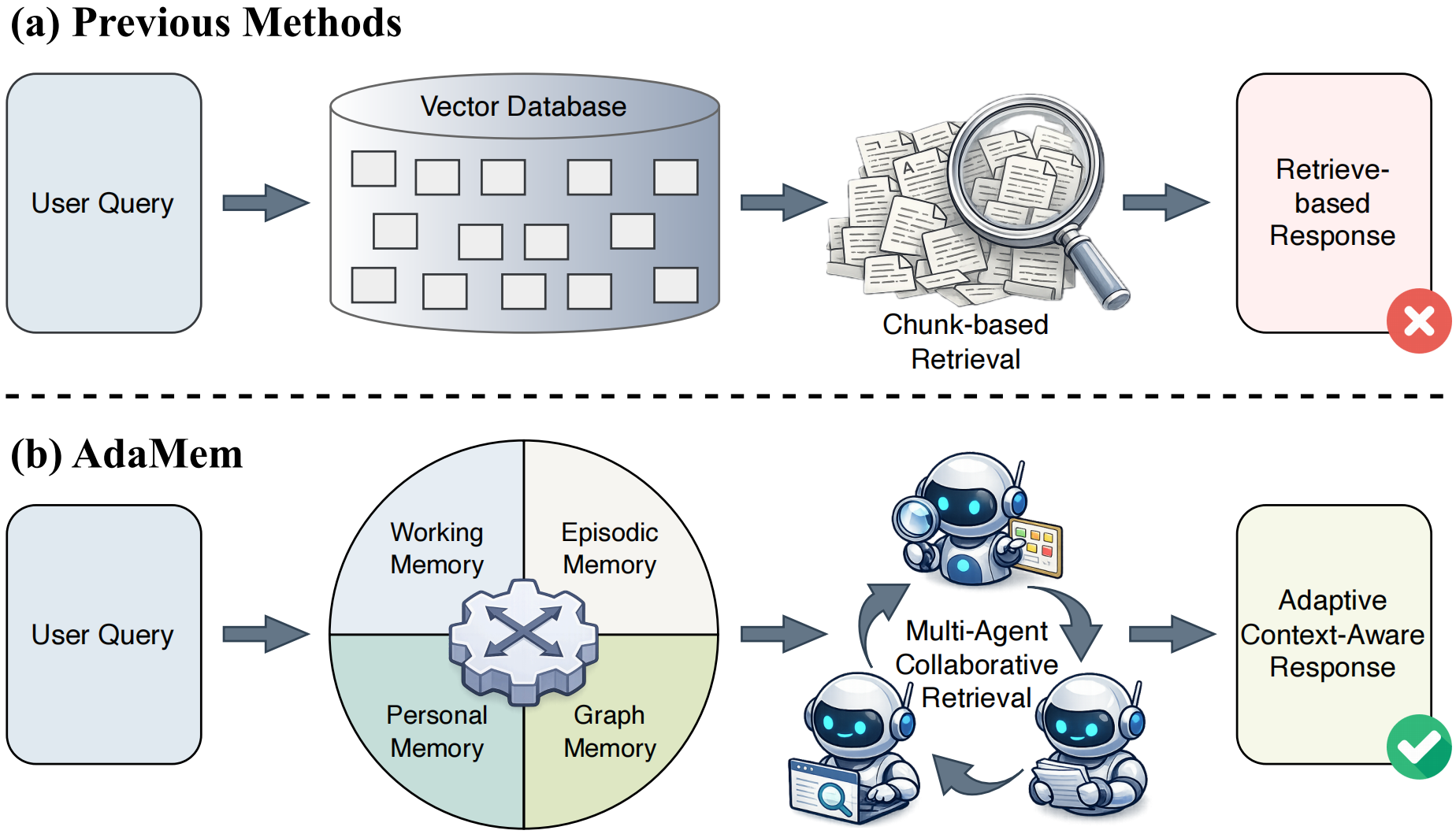}
\caption{{{\bf Comparison of previous methods and AdaMem. }Conventional approaches rely on fixed-length chunks or coarse summaries with semantic retrieval, while AdaMem emphasizes user-centric adaptive structured memories and multi-agent collaborative retrieval. } 
}
\label{fig:teaser}
\vspace{-0.1in}
\end{figure}

Many recent agent frameworks therefore augment LLMs with explicit external memory modules that support incremental writing, updating, and retrieval throughout interaction~\citep{yan2025memory, zep}. Despite this progress, current approaches still face three limitations.

\paragraph{Limitation 1:} Memory systems that rely primarily on semantic retrieval may overlook evidence that is not lexically or semantically similar to the query, but is still crucial for user-centric understanding, such as stable preferences, personal attributes, or broader behavioral patterns.

\paragraph{Limitation 2:} When related experiences are stored as isolated fragments, their temporal and causal coherence can be weakened, making it difficult to reconstruct how events unfolded and how different pieces of evidence should be connected during reasoning.

\paragraph{Limitation 3:} Different questions require different memory structures and retrieval strategies. As illustrated in Figure~\ref{fig:teaser}, many systems construct memory entries using either fixed-length text chunks~\citep{zhang2025survey,wu2025human}. Such static segmentation is often a poor fit for long-horizon reasoning: overly coarse memories may introduce substantial irrelevant context, while overly fine-grained fragments can obscure dependencies across events and topics.

These limitations suggest that effective long-horizon memory should be both \textit{structured} and \textit{adaptive}: it should preserve information at multiple levels of abstraction while dynamically selecting retrieval routes that match each question. Motivated by this observation, we propose \( \mathit{AdaMem}\), an adaptive user-centric memory framework for long-horizon dialogue agents. AdaMem maintains participant-specific working, episodic, persona, and graph memories, organizing evidence around the user and the assistant in a unified yet target-aware manner. Unlike prior systems that rely on a monolithic controller for memory writing, retrieval, verification, and response generation, AdaMem uses a role-specialized agent pipeline. The Memory Agent maintains structured memories, the Research Agent performs question-conditioned retrieval and evidence integration with reflection, and the Working Agent turns the collected evidence into a concise answer. This decomposition reduces interference between memory maintenance and answer-time reasoning, enabling finer control over retrieval, verification, and memory evolution. Together, these design choices recover user information beyond pure semantic similarity, preserve links across related events, and adapt retrieval granularity to the demands of the question.
In summary, our contributions are:
\begin{itemize}
    \item We introduce AdaMem, an adaptive user-centric memory framework for long-horizon dialogue agents that organizes dialogue history into complementary working, episodic, persona, and graph-based memory structures.
    \item We propose a question-conditioned retrieval and response pipeline that resolves target participants, invokes relation-aware graph expansion only when needed, and uses specialized agents for evidence synthesis and answer generation.
    \item We validate our approach on the LoCoMo and PERSONAMEM benchmarks, achieving state-of-the-art performance and demonstrating its strong effectiveness.
\end{itemize}

\section{Related Works}

\subsection{Agentic Memory}

Recent research on memory systems for large language model agents has evolved from simple context extension toward more structured and adaptive management. Early approaches typically process long contexts by partitioning them into smaller chunks \citep{memorybank, aios, scm, agentlite}. Subsequent work introduces more advanced memory mechanisms. For instance, MemGPT \citep{memgpt} manages long-term memory through paging and segmentation. Later studies explore more modular and system-level designs; Mem0 \citep{mem0} abstracts memory as an independent layer for long-term management. Meanwhile, several works investigate structured memory representations to improve organization, including graph-based memories such as A-Mem \citep{amem} and semantic memory structures built upon events or temporal knowledge graphs, such as Zep \citep{zep}. Despite these advances, many existing approaches still rely on static retrieval strategies or largely unstructured storage, which may lead to information fragmentation and limited coordination across different abstraction levels and task categories. Designing a unified and adaptive memory system that robustly supports long-term interactions therefore remains an important challenge.

\subsection{Multi-Agent LLMs}

Recent studies have shown that multi-agent LLMs can effectively address complex tasks by enabling role specialization, collaborative problem solving, and interactive decision making \citep{tran2025multi, cemri2025multi, hammond2025multi, masrouter, zhang2025agent}. At the same time, an increasing number of works on agentic memory aim to enhance the capability of large language model agents to model and retain long-term information \citep{yan2025memory, wang2025mem}. Although these efforts provide useful perspectives on coordination and task decomposition, they typically do not focus on how long-horizon dialogue evidence should be organized and adaptively retrieved in a user-centric manner. A recent study, MIRIX \citep{wang2025mirix}, takes an initial step in this direction by introducing specialized agents for memory organization. However, it does not provide explicit mechanisms to ensure the consistency of long-term memory. Motivated by these observations, our work draws on role specialization from the multi-agent literature, but focuses primarily on user-centric memory construction and question-conditioned retrieval for long-horizon dialogue agents.

\section{Approach}
\label{sec:approach}

\begin{figure*}
\centering
\includegraphics[width=1.0\textwidth]{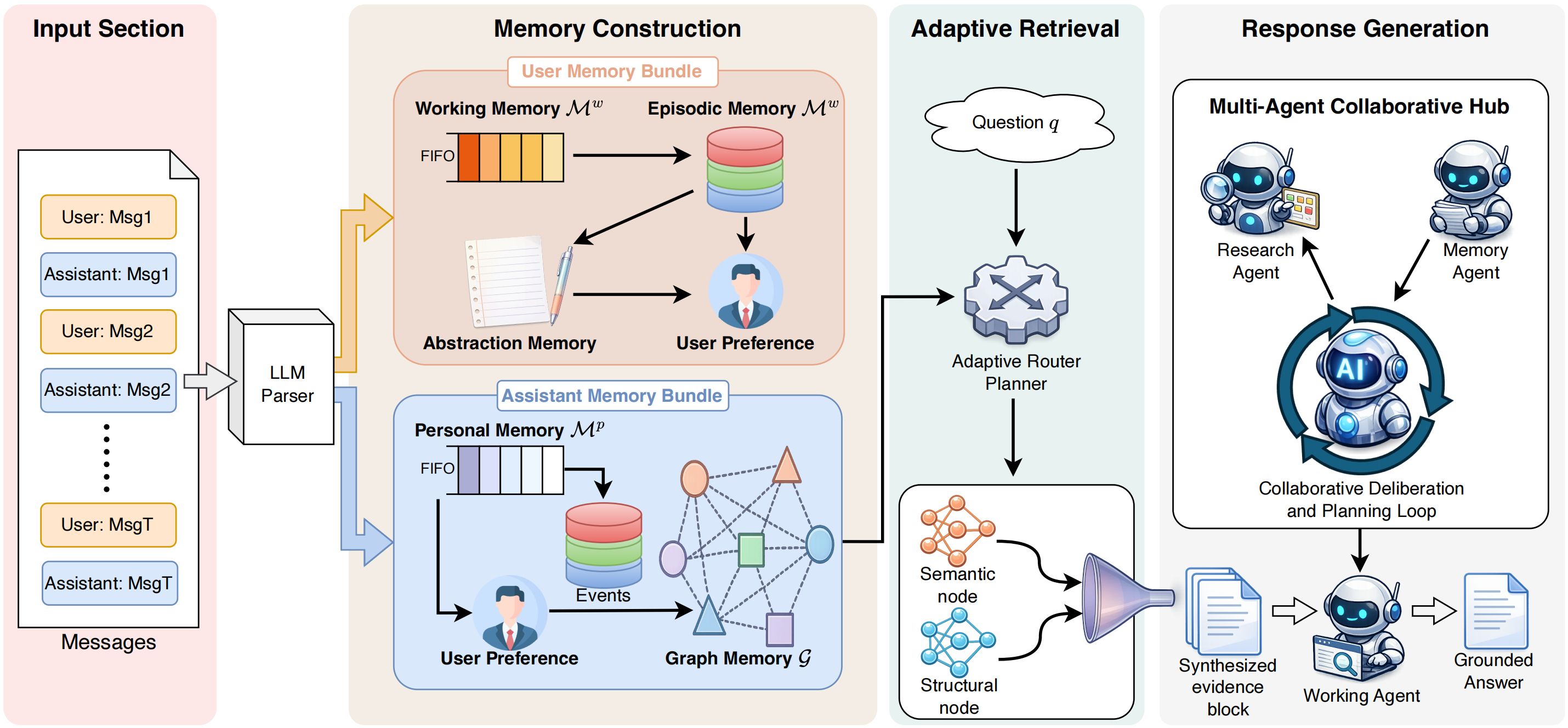}
\caption{
\textbf{Model overview.} Dialogue history is organized into working, episodic, persona, and graph memories, and question answering is performed through target-aware, question-conditioned retrieval and role-specialized evidence synthesis.
}
\label{fig:overview}
\vspace{-0.1in}
\end{figure*}

We present AdaMem, a memory-augmented dialogue framework that continuously organizes user-centric memories at inference time and answers questions through adaptive retrieval over heterogeneous memory sources. AdaMem consists of four tightly coupled components: memory construction, question-conditioned retrieval planning, evidence fusion, and response generation. The overall pipeline is illustrated in Figure~\ref{fig:overview}.

\subsection{Method Overview}
Given a dialogue history $\mathcal{D}=\{u_t\}_{t=1}^{T}$ and a question $q$, AdaMem returns an answer $a$ grounded in dynamic memories. AdaMem maintains participant-specific memory bundles for the user and the assistant, and each bundle contains four memory structures:
\begin{itemize}
    \item \textbf{Working Memory} $\mathcal{M}^{\mathrm{w}}$: a bounded FIFO buffer that preserves recent conversational context and short-term discourse states.
    \item \textbf{Episodic Memory} $\mathcal{M}^{\mathrm{e}}$: long-term structured records including events, facts, attributes, and topic-centric summaries.
    \item \textbf{Persona Memory} $\mathcal{M}^{\mathrm{p}}$: compact user profiles distilled from episodic evidence to capture relatively stable preferences and traits.
        \item \textbf{Graph Memory} $\mathcal{G}$: a heterogeneous graph connecting messages, topics, facts, attributes, and event or persona snapshots for relation-aware retrieval. Detailed graph construction rules are provided in Appendix~\ref{app:graph}.
\end{itemize}
This participant-specific organization is important because many questions in multi-party conversations implicitly target one speaker, both speakers, or an ambiguous referent. Accordingly, AdaMem follows a single end-to-end pipeline: it first writes each utterance into structured participant memories, then resolves the likely target participant for $q$, retrieves evidence from one or more memory channels, and finally produces the answer from the fused evidence set $\mathcal{R}(q)$.

\subsection{Memory Construction}
\paragraph{Message understanding and normalized write.}
For each incoming utterance $u_t$ from participant $p$, the Memory Agent first produces a normalized record $z_t$ containing a short summary, topic, attitude, reason, factual snippets, attributes, timestamp, and speaker identity. All downstream memory updates operate on $z_t$ rather than on raw free-form text, so the same canonical parse is reused across working, episodic, persona, and graph memories. This design reduces prompt drift across memory modules and makes the write path explicit.

\paragraph{Working-to-episodic consolidation.}
For each participant, working memory is a bounded FIFO queue with capacity $C_{\mathrm{w}}$. When the queue becomes full, AdaMem pops the oldest contiguous segment of $r$ messages and consolidates that segment into $\mathcal{M}^{\mathrm{e}}$. The segment is therefore determined by recency order rather than by question-aware salience, preventing future questions from implicitly influencing write-time memory selection. Within the popped segment, three router modules independently process event-, fact-, and attribute-level evidence. Each router predicts one of \texttt{ADD}, \texttt{UPDATE}, or \texttt{IGNORE}, together with a target key if an existing entry should be revised. The resulting updates populate event, fact, and attribute stores, while the original messages remain available as cacheable provenance for later evidence recovery.

\paragraph{Topic regrouping, persona refresh, and graph synchronization.}
After consolidation, AdaMem converts fine-grained episodic keys into reusable higher-level memories in two stages. First, event or attribute keys are embedded and linked by a sparse nearest-neighbor graph: each key is connected to its most similar peer, and the connected components define merge groups. Second, an LLM merge prompt rewrites each group into a topic-centric or aspect-centric summary. Topic groups are used to build topic episodic memories and preference-oriented persona descriptors, while clustered attributes are merged into aspect-based persona summaries. In parallel, both message-level and consolidated records are indexed into $\mathcal{G}$, so later retrieval can move between local discourse evidence and longer-term structured abstractions.

\subsection{Question-Conditioned Retrieval}
\paragraph{Target participant resolution.}
Before retrieval, AdaMem resolves whether $q$ refers to the user, the assistant, both, or an ambiguous participant. The implementation uses a lightweight four-way resolver based on explicit participant mentions; if the referent is ambiguous, the system avoids a hard commitment and instead runs a small retrieval on both participant bundles before fusion. This makes target uncertainty visible to the retrieval stage rather than hiding it inside later answer generation.

\paragraph{Route planning.}
Given $q$, AdaMem builds a question-conditioned route plan $\pi(q)$ that specifies whether graph expansion should be used, how far it may propagate, how many graph seeds are activated, which edge-type priors are applied, and how baseline and graph evidence will be fused. The planner first applies deterministic cue detection over temporal, relational, attribute, and single-hop question patterns. These cues initialize a rule-based plan. When the question remains uncertain, an optional LLM refinement step can revise the plan, but only within narrow bounded ranges so that the planner remains conservative and semantic retrieval stays dominant for simple questions. As a result, single-hop factoid questions usually remain on lightweight semantic retrieval, whereas temporal or causal questions trigger broader structural exploration. 

\paragraph{Target-aware baseline retrieval.}
For a selected participant bundle, baseline retrieval aggregates semantic candidates from persona summaries, episodic facts, and topic-linked messages:
\begin{equation}
\mathcal{R}_{\mathrm{base}}(q)=\mathrm{TopK}\!\left(
\mathcal{R}_{\mathrm{attr}}(q)\cup
\mathcal{R}_{\mathrm{fact}}(q)\cup
\mathcal{R}_{\mathrm{topic}}(q)
\right),
\end{equation}
where attribute candidates come from persona summaries, fact candidates come from episodic fact memory, and topic candidates are linked back to original messages through topic-to-message maps. Beyond pure top-$K$ aggregation, AdaMem applies two recovery mechanisms. First, high-confidence fact hits reactivate their supporting discourse messages. Instead of attaching all neighboring contexts, AdaMem uses a score-drop rule over ranked fact matches to decide how many supporting messages should be reintroduced. Second, a lightweight keyword backoff over a word-to-detail index recalls raw messages that are weakly covered by dense retrieval but lexically salient for the question.

\paragraph{Graph retrieval and evidence fusion.}
When $\pi(q)$ requests relation-aware evidence, AdaMem selects top semantic seed nodes in $\mathcal{G}$ and performs bounded multi-hop expansion. If a seed node $u$ reaches a neighbor $v$ through edge type $e$, the propagated score is updated by a fixed multiplicative rule
\begin{equation}
 s_{d+1}(v)=s_d(u)\cdot w_e \cdot \lambda,
\end{equation}
where $w_e$ is the edge-type prior and $\lambda$ is a hop-decay factor. Owner-aware filtering is applied only when the target participant is unambiguous. This yields $\mathcal{R}_{\mathrm{graph}}(q)$, which is fused with baseline evidence by
\begin{equation}
\begin{split}
\mathrm{score}(m\mid q) =
\alpha\,s_{\mathrm{base}}(m,q)
+ \beta\,s_{\mathrm{graph}}(m,q) \\
+ \gamma\,s_{\mathrm{recency}}(m)
+ \delta\,s_{\mathrm{fact}}(m).
\end{split}
\end{equation}
Here $s_{\mathrm{base}}(m,q)$ and $s_{\mathrm{graph}}(m,q)$ are rank-derived scores from baseline and graph retrieval, respectively; $s_{\mathrm{recency}}(m)$ is a linear recency prior over the merged candidate list; and $s_{\mathrm{fact}}(m)$ is a small confidence bonus for items also supported by fact retrieval. Notably, for the fusion weights, a global default prior is fixed before evaluation.

\subsection{Multi-Agent Collaboration}
\paragraph{Memory Agent.}
The Memory Agent is responsible for online message understanding and memory updates. For each new utterance, it extracts a normalized representation, writes it into the participant-specific working memory, triggers working-to-episodic consolidation when the short-term buffer saturates, and synchronizes the resulting message and memory artifacts to graph memory. Persona descriptors are then refreshed from aggregated episodic evidence at the indexing stage, so the system maintains both up-to-date local context and compact long-term user models.

\paragraph{Research Agent.}
At question answering time, the Research Agent performs iterative evidence gathering over the unified retrieval interface described above. It follows a \textit{Planning $\rightarrow$ Search $\rightarrow$ Integrate $\rightarrow$ Reflection} loop: it first decomposes the information needed to answer $q$, then issues one or more retrieval requests through the same participant-aware retrieval API, integrates newly recovered evidence into a consolidated research summary, and decides whether additional search is still necessary. Importantly, this agent-level planning is distinct from the route plan $\pi(q)$: the Research Agent decides \emph{what} missing information to ask for, while the route planner described above decides \emph{how} each retrieval call is executed.

\paragraph{Working Agent.}
The Working Agent converts the research summary into the final concise answer. It conditions generation primarily on the integrated summary returned by the Research Agent and, when needed, supplements it with high-confidence persona attributes or factual snippets as auxiliary grounding. This separation allows evidence collection and answer realization to be optimized for different roles while preserving a single memory interface.

AdaMem answers each question through a fixed collaboration order among these roles. As a result, response generation remains tightly coupled with the same user-centric memory interface and retrieval backbone introduced above, while allocating explicit deliberation to multi-step evidence synthesis before final answer generation.


\section{Experiments}
\subsection{Experimental Setting}

\paragraph{Benchmarks.}
To assess the effectiveness of our approach, we conduct experiments on the LoCoMo benchmark \citep{maharana2024evaluating}. LoCoMo poses a challenging setting for long-context modeling, consisting of dialogue histories that span an average of 35 sessions and roughly 9,000 tokens. Following the benchmark’s standard evaluation protocol, we report quantitative results across four core capabilities: single-hop reasoning, multi-hop reasoning, temporal reasoning, and open-domain question answering. The original benchmark also includes an adversarial question category designed to test a model’s ability to identify unanswerable queries. 

To further examine the generalization capability of our approach, we conduct additional experiments on the PERSONAMEM benchmark \citep{jiang2025know}. It is designed to assess how well large language models maintain and update user representations and produce personalized responses over extended interactions. The benchmark comprises multi-session dialogue histories in which user attributes and preferences gradually evolve as a result of life events and changing contexts. Following its standard evaluation protocol, we report quantitative results across seven categories, each targeting different aspects of user modeling and memory utilization.

\paragraph{Evaluation Metrics.}
For the LoCoMo benchmark, we follow the standard evaluation protocol and report F1 and BLEU-1 as the primary metrics. For the PERSONAMEM benchmark, which primarily consists of multiple-choice questions, we evaluate model performance using accuracy.

\subsection{Implementation Details}

We conduct experiments on both closed-source APIs and open-source models. The closed-source models include GPT-4.1-mini and GPT-4o-mini \citep{gpt4}. For open-source models, we evaluate Qwen3-4B-Instruct 
and Qwen3-30B-A3B-Instruct \citep{qwen3}. 
Experiments are conducted on a NVIDIA RTX A800 GPU. 
We ensure that the AdaMem framework uses the same backbone model as the response generator. To ensure reproducibility, the temperature is fixed to 0 in all experiments. For the RAG component in our framework, the retrieval top-$k$ is set to 10, and the maximum number of retrieval iterations $L_i$ is limited to 2 for efficiency. All memory embeddings are computed using the \texttt{all-MiniLM-L6-v2} model \citep{reimers2019sentence}. More implementation details are provided in Appendix \ref{app:more_impl} and prompt templates are provided in Appendix \ref{app:prompt}.

\subsection{Comparison to competitive approaches}

\paragraph{Baselines.}
We compare AdaMem with five representative open-source memory frameworks: (1) \textit{MemGPT} \citep{memgpt}, a framework that addresses long-context limitations through an OS-inspired memory management mechanism; (2) \textit{A-Mem} \citep{amem}, an agentic memory system that enables dynamic organization and evolution of long-term memory; (3) \textit{Mem0} \citep{mem0}, a memory-centric architecture that provides scalable long-term memory; (4) \textit{LangMem} \citep{langmem}, a framework that explicitly models long-term memory; (5) \textit{Zep} \citep{zep}, a memory layer that represents conversational memory using a temporally aware knowledge graph.

\paragraph{Results on LoCoMo.}

The quantitative results on the LoCoMo benchmark using closed-source backbones are summarized in Table~\ref{tab:main_locomo}. With GPT-4.1-mini, AdaMem achieves an overall F1 score of 44.65\%, corresponding to a +4.4\% relative improvement over the previous state-of-the-art method. The improvement is consistent across evaluation metrics, with the largest gain observed in the temporal question category, where AdaMem improves the F1 score by up to +23.4\%. Using GPT-4o-mini, AdaMem achieves an overall F1 score of 41.84\%, which corresponds to a larger relative improvement of +12.8\% over the previous state-of-the-art method. AdaMem does not outperform the strongest baselines in the Open Domain categories, likely because some tasks favor methods tailored to specific structural assumptions. This result suggests a trade-off between optimizing for specialized task structures and achieving consistent performance across diverse long-horizon interactions. Overall, the results demonstrate the effectiveness and robustness of our approach across different closed-source backbones. We further report results using open-source models in Table~\ref{tab:abl_model_size}. Case studies are provided in Appendix~\ref{app:case}.

\begin{table*}[t]
\centering
\caption{\textbf{Performance on the LoCoMo benchmark.} The best performance is highlighted in \textbf{bold}, and the second-best is \underline{underlined}.}
\label{tab:main_locomo}
\resizebox{1\textwidth}{!}{%
\begin{tabular}{l|l|cc|cc|cc|cc|cc}
\toprule
\multirow{2}{*}{\textbf{Model}} & \multirow{2}{*}{\textbf{Method}} & \multicolumn{2}{c|}{\textbf{Multi-hop}} & \multicolumn{2}{c|}{\textbf{Temporal}} & \multicolumn{2}{c|}{\textbf{Open Domain}} & \multicolumn{2}{c|}{\textbf{Single-hop}} & \multicolumn{2}{c}{\textbf{Overall}} \\
 & & \textbf{F1} & \textbf{BLEU-1} & \textbf{F1} & \textbf{BLEU-1} & \textbf{F1} & \textbf{BLEU-1} & \textbf{F1} & \textbf{BLEU-1} & \textbf{F1} & \textbf{BLEU-1}  \\
\midrule

\multirow{6}{*}{\textbf{GPT-4.1-mini}} 
 & MemGPT & 10.35 & 10.26 & 30.06 & 24.03 & 28.35 & \underline{23.49} & 22.95 & 17.10 & 22.50 & 17.64 \\
 & A-Mem & 11.16 & 11.07 & 42.30 & 33.75 & \textbf{30.69} & \textbf{25.38} & 24.84 & 18.54 & 26.37 & 20.70 \\
 & Mem0 & 35.19 & 26.55 & 34.38 & 29.07 & 20.88 & 15.57 & 42.66 & 36.90 & 38.16 & 32.04 \\
 & LangMem & \underline{36.45} & \underline{28.53} & \underline{42.57} & \underline{35.91} & \underline{28.80} & 23.13 & \underline{44.82} & \underline{38.25} & \underline{41.76} & \underline{35.10} \\
 & Zep & 26.73 & 17.91 & 20.97 & 17.55 & 21.24 & 16.92 & 39.96 & 35.10 & 32.40 & 27.09 \\
 \rowcolor{cyan!10} & \textbf{AdaMem} & \textbf{37.70} & \textbf{32.08} & \textbf{55.90} & \textbf{42.37} & 25.87 & 21.70 & \textbf{44.84} & \textbf{40.02} & \textbf{44.65} & \textbf{37.92} \\
\midrule

\multirow{6}{*}{\textbf{GPT-4o-mini}} 
 & MemGPT & 10.08 & 9.99 & 29.34 & 23.49 & \underline{27.72} & \underline{22.95} & 22.41 & 16.74 & 21.96 & 17.28 \\
 & A-Mem & 10.89 & 10.80 & \underline{41.31} & 33.03 & \textbf{29.97} & \textbf{24.84} & 24.30 & 18.09 & 25.74 & 22.95 \\
 & Mem0 & \underline{30.60} & \underline{22.50} & 39.60 & 33.57 & 24.21 & 17.28 & \underline{39.69} & \underline{30.60} & \underline{37.08} & \underline{30.51} \\
 & LangMem & 30.24 & 21.60 & 28.89 & 23.67 & 26.55 & 21.24 & 35.01 & 29.88 & 32.31 & 26.55 \\
 & Zep & 24.57 & 17.28 & 39.96 & \underline{34.02} & 20.43 & 14.04 & 35.46 & 30.06 & 33.48 & 27.63 \\
 \rowcolor{cyan!10} & \textbf{AdaMem} & \textbf{35.18} & \textbf{26.32} & \textbf{51.49} & \textbf{37.06} & 25.82 & 21.46 & \textbf{42.21} & \textbf{36.43} & \textbf{41.84} & \textbf{33.78} \\

\bottomrule
\end{tabular}
}
\end{table*}

\paragraph{Results on PERSONAMEM.}

To evaluate the generalization capability of AdaMem across benchmarks, we further evaluate it on the PERSONAMEM benchmark. As shown in Table~\ref{tab:main_personamem}, AdaMem achieves 63.25\% accuracy, outperforming all baselines with a relative improvement of 5.9\%. Notably, AdaMem demonstrates a substantial advantage on the generalize to new scenarios task, achieving a relative improvement of 27.3\%. These consistent gains across benchmarks with different data distributions suggest that AdaMem generalizes effectively to diverse long-term reasoning scenarios.

\begin{table*}[t]
\centering
\caption{\textbf{Performance on the PERSONAMEM benchmark.}}
\label{tab:main_personamem}
\small
\setlength{\tabcolsep}{3pt}
\begin{tabular}{l c c c c c c c}
\toprule
\textbf{Method} &
\begin{tabular}[c]{@{}c@{}}Recall user\\ shared facts\end{tabular} &
\begin{tabular}[c]{@{}c@{}}Suggest new\\ ideas\end{tabular} &
\begin{tabular}[c]{@{}c@{}}Track full\\ preference\\ evolution\end{tabular} &
\begin{tabular}[c]{@{}c@{}}Revisit reasons\\ behind\\ preference updates\end{tabular} &
\begin{tabular}[c]{@{}c@{}}Provide preference-\\aligned\\ recommendations\end{tabular} &
\begin{tabular}[c]{@{}c@{}}Generalize to\\ new scenarios\end{tabular} &
\textbf{Average} \\
\midrule
A-Mem    & 63.01 & \textbf{27.96} & \underline{54.68} & \underline{85.86} & \textbf{69.09} & \underline{57.89} & \underline{59.75} \\
Mem0     & 32.13 & 15.05 & \underline{54.68} & 80.81 & 52.73 & \underline{57.89} & 48.55 \\
LangMem   & 31.29 & \underline{24.73} & 53.24 & 81.82 & 40.00 &  8.77 & 40.64 \\
\rowcolor{cyan!10}
\textbf{AdaMem} & \textbf{67.81} & 21.51 & \textbf{61.15} & \textbf{89.90} & \underline{65.45} & \textbf{73.68} & \textbf{63.25} \\
\bottomrule
\end{tabular}
\label{tab:personamem_llm}
\vspace{-0.1in}
\end{table*}

\subsection{Ablation Study} 

Unless otherwise stated, we perform all ablations of AdaMem on the LoCoMo benchmark using the GPT-4.1-mini backbone.

\paragraph{Components.}

We perform component ablations to quantify the contribution of AdaMem's main design choices. As shown in Table~\ref{tab:abl_component}, removing any one of the three components consistently reduces performance, suggesting that the improvement of AdaMem does not come from a single dominant module. Disabling graph memory produces the largest drop, reducing overall F1 from 44.65 to 42.63. This result is consistent with our motivation in Section~\ref{sec:approach}: relation-aware memory is important for recovering cross-turn dependencies and temporally linked evidence that may be missed by semantic retrieval alone. Removing the fusion module also leads to clear degradation (42.77 F1), indicating that jointly combining baseline retrieval, graph evidence, and lightweight temporal/factual signals yields more reliable evidence selection than relying on a single signal source. Replacing the multi-agent response pipeline with a single-agent variant causes a smaller but still consistent decline (43.24 F1), suggesting that role specialization mainly improves evidence organization and final answer synthesis after retrieval. Overall, the three components are complementary: graph memory improves structural evidence coverage, fusion improves evidence aggregation, and multi-agent coordination improves downstream reasoning quality.

\begin{table}[t]
\begin{center}
\caption{\textbf{Ablation on key components.} }
\label{tab:abl_component}
\resizebox{1\linewidth}{!}{
\begin{tabular}{l c c c c c}
\toprule
\textbf{Configuration} & \textbf{Graph} & \textbf{Fusion} & \textbf{Multi-agent} & \textbf{F1} & \textbf{BLEU-1} \\
\midrule
\rowcolor{cyan!10}
\textbf{AdaMem (full)} & \cmark & \cmark & \cmark & \textbf{44.65} & \textbf{37.92}  \\
w/o graph & \xmark & \cmark & \cmark & 42.63 & 35.85  \\
w/o fusion & \cmark & \xmark & \cmark & 42.77 & 36.26 \\
w/o multi-agent & \cmark & \cmark & \xmark & 43.24 & 36.34  \\
\bottomrule
\end{tabular}
}
\end{center}
\vspace{-0.1in}
\end{table}


\paragraph{Model Sizes.} 

We further study whether AdaMem remains effective when the response backbone changes in both scale and model family. Table~\ref{tab:abl_model_size} compares two closed-source GPT-4 backbones with two open-source Qwen3 models while keeping the AdaMem framework unchanged. 
AdaMem generalizes well across backbone families. Even with the smaller open-source Qwen3-4B-Instruct model, AdaMem still reaches 36.78 F1 on LoCoMo, showing that the memory architecture remains useful under a constrained model budget. Scaling the backbone to Qwen3-30B-A3B-Instruct yields consistent gains on every category, improving overall performance by +6.24 F1 over Qwen3-4B. The largest gain appears on temporal reasoning (+13.17 F1), followed by multi-hop reasoning (+6.91 F1), which suggests that larger models make better use of AdaMem's structured evidence when questions require linking events over time or synthesizing multiple memory fragments. Overall, this experiment shows that AdaMem is both robust under smaller models and able to translate additional model capacity into stronger long-horizon reasoning.

\begin{table*}[t]
\centering
\caption{\textbf{Ablation on model sizes.}}
\label{tab:abl_model_size}
\resizebox{1\textwidth}{!}{%
\begin{tabular}{l|cc|cc|cc|cc|cc}
\toprule
\multirow{2}{*}{\textbf{Model}} &  \multicolumn{2}{c|}{\textbf{Multi-hop}} & \multicolumn{2}{c|}{\textbf{Temporal}} & \multicolumn{2}{c|}{\textbf{Open Domain}} & \multicolumn{2}{c|}{\textbf{Single-hop}} & \multicolumn{2}{c}{\textbf{Overall}} \\
 & \textbf{F1} & \textbf{BLEU-1} & \textbf{F1} & \textbf{BLEU-1} & \textbf{F1} & \textbf{BLEU-1} & \textbf{F1} & \textbf{BLEU-1} & \textbf{F1} & \textbf{BLEU-1}  \\
\midrule

\textbf{GPT-4o-mini} & 35.18 & 26.32 & 51.49 & 37.06 & 25.82 & 21.46 & 42.21 & 36.43 & 41.84 & 33.78 \\

\textbf{GPT-4.1-mini} & 37.70 & 32.08 & 55.90 & 42.37 & 25.87 & 21.70 & 44.84 & 40.02 & 44.65 & 37.92 \\

\midrule
\textbf{Qwen3-4B-Instruct} & 29.50 & 23.35 & 40.45 & 29.74 & 20.05 & 17.68 & 39.74 & 34.98 & 36.78 & 30.68 \\

\textbf{Qwen3-30B-A3B-Instruct} & 30.82 & 25.17 & 43.51 & 30.90 & 26.55 & 23.31 & 40.68 & 35.99 & 38.58 & 32.16 \\


\bottomrule
\end{tabular}
}
\vspace{-0.1in}
\end{table*}

\paragraph{Hyperparameters.}

\begin{figure}
\centering
\includegraphics[width=0.48\textwidth]{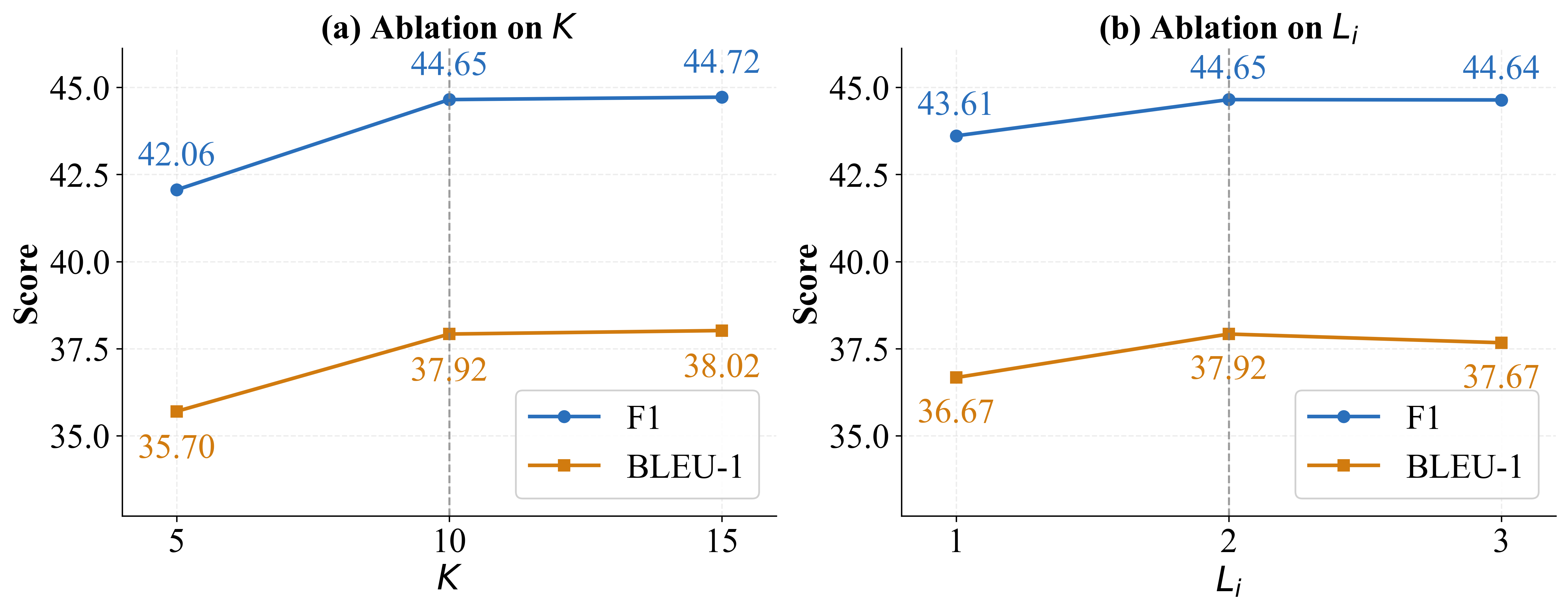}
\caption{{\bf Ablation on key hyperparameters. } 
}
\label{fig:abl_para}
\vspace{-0.1in}
\end{figure}

We study the sensitivity of AdaMem to two key hyperparameters that directly control evidence breadth and deliberation depth: the retrieval top-$K$ and the maximum number of Research Agent iterations $L_i$. Results are shown in Figure~\ref{fig:abl_para}. This analysis helps determine whether AdaMem's gains come from a stable operating regime or from overly aggressive evidence accumulation.

 Fixing $L_i=2$, we vary $K \in \{5,10,15\}$. Performance improves substantially from $K=5$ to $K=10$, indicating that very small candidate pools often fail to cover the dispersed evidence needed for multi-session reasoning. Increasing $K$ further to 15 yields only marginal gains over $K=10$, suggesting diminishing returns once the main supporting evidence has already been retrieved. We therefore use $K=10$ as the default setting, as it achieves nearly the best performance while avoiding the additional latency and noise introduced by a larger evidence set.

 Fixing $K=10$, we vary $L_i \in \{1,2,3\}$. Both F1 and BLEU-1 peak at $L_i=2$. Using only 1 iteration is often insufficient for questions that require decomposition or follow-up retrieval, whereas extending the loop to 3 iterations slightly hurts performance, likely because later rounds accumulate redundant or weakly related evidence. This trend is consistent with the design of the Research Agent: iterative retrieval is beneficial, but only under a tight budget that preserves evidence precision. Thus, we use $L_i=2$ as the default setting. 
 
 Overall, AdaMem is moderately sensitive rather than brittle, and the default configuration used in our main experiments provides a favorable performance--efficiency trade-off.

\subsection{Efficiency Analysis}

We analyze the performance-efficiency trade-off of AdaMem. Table~\ref{tab:efficiency} reports the average input token budget, end-to-end inference latency, and overall F1 and BLEU-1. AdaMem is not the cheapest system in raw computation: Mem0 uses fewer input tokens and lower latency. However, AdaMem delivers the strongest answer quality, reaching 44.65 F1, which corresponds to absolute gains of +7.57 F1 over Mem0. Compared with A-Mem and Zep, AdaMem also achieves substantially better accuracy while operating with a token budget of the same order.

These results suggest that AdaMem's advantage does not come from aggressively minimizing retrieval cost, but from allocating a moderate amount of additional computation to recover higher-quality evidence. The extra latency is consistent with our design: question-conditioned route planning, graph-based expansion, and the role-specialized response loop introduce overhead, but they also improve evidence coverage and synthesis for long-horizon questions. Overall, AdaMem occupies a favorable operating point among memory-based methods, converting a moderate increase in token usage and latency into clearly stronger reasoning performance.

\begin{table}[t]
\centering
\caption{\textbf{Performance-efficiency trade-off.}}
\label{tab:efficiency}
\resizebox{1\linewidth}{!}{%
\begin{tabular}{ccccc}
\toprule
\textbf{Method} &\textbf{F1}  & \textbf{BLEU-1} & \textbf{Tokens} & \textbf{Latency (s)} \\
\midrule
A-Mem & 26.37 & 20.70 & 2720 & 3.227  \\
Mem0 & 38.16 & 32.04 & 1340 & 3.739   \\
Zep & 32.40 & 27.09 & 2461 & 3.255 \\
\rowcolor{cyan!10}
\textbf{AdaMem} & 44.65 & 37.92 & 2248 & 4.722  \\ 
\bottomrule
\end{tabular}%
}
\vspace{-0.1in}
\end{table}

\section{Conclusion}
\label{sec:bibtex}

In this paper, we presented AdaMem, an adaptive user-centric memory framework for long-horizon dialogue agents. AdaMem combines participant-specific working, episodic, persona, and graph-based memories with question-conditioned retrieval planning and unified evidence fusion, enabling the system to retrieve and integrate evidence in a more structured and target-aware manner. Experiments on long-horizon reasoning and user modeling benchmarks demonstrate the effectiveness of our framework. In particular, AdaMem achieves state-of-the-art performance on LoCoMo, showing the value of adaptive memory organization and retrieval for complex multi-session interactions. More broadly, our results suggest that memory for long-horizon dialogue agents should move beyond uniform storage and fixed retrieval heuristics toward more adaptive, question-aware, and user-centric designs.

\section*{Limitations}

AdaMem improves answer quality through structured memories, adaptive retrieval, and role-specialized evidence synthesis, but this design also increases system complexity, token cost, and latency. In addition, the framework still depends on upstream parsing and backbone reasoning, making errors in target resolution, entity linking, and temporal normalization difficult to recover. In future work, we intend to explore solutions to these challenges in order to further enhance both the efficiency and the generality of the proposed framework.



\bibliography{custom}

\clearpage
\appendix

\section{Organization} \label{app:org}

The appendix contains the following sections: 
\begin{itemize}
	\item More implementation details are provided in Section~\ref{app:more_impl}.  

    \item Graph construction details are provided in Section~\ref{app:graph}.

    \item Case studies are provided in Section~\ref{app:case}.

    \item Prompt templates are provided in Section~\ref{app:prompt}.
	
\end{itemize}






\section{More Implementation Details}
\label{app:more_impl}

This section documents the concrete decision rules used in our implementation so that AdaMem's gains can be attributed to its retrieval design rather than to hidden prompt engineering or benchmark-specific tuning. Unless otherwise stated, the benchmark settings below are fixed before evaluation and reused across runs. The repository also contains looser demo defaults for interactive examples; the paper numbers should therefore be interpreted with the benchmark configuration rather than the demo configuration.

\subsection{Target Participant Resolution}
AdaMem uses a lightweight four-way target resolver. If a question explicitly mentions the user name, the target is the user; if it explicitly mentions the assistant name, the target is the assistant; if it mentions both names, the target is \texttt{both}; otherwise the target is \texttt{ambiguous}. For \texttt{both} and \texttt{ambiguous}, the system does not force an early commitment. Instead, it performs a small retrieval on both participant bundles and merges the candidates before final reranking. In graph retrieval, owner-aware filtering is applied only when the target is unambiguous. We do not use a separately trained target-participant classifier.

\subsection{Route Planning and Graph Expansion}
The route planner first applies deterministic cue detection:
\begin{itemize}
    \item temporal cues: ``when'', ``date'', ``year'', ``month'', ``time'', ``last'', ``ago'', ``before'', ``after'';
    \item relation cues: ``why'', ``because'', ``cause'', ``how'', ``relationship'', ``connect'', ``between'';
    \item attribute cues: ``prefer'', ``like'', ``favorite'', ``personality'', ``trait'', ``attribute'';
    \item single-hop cues: ``who'', ``what'', ``where'', ``which'', ``name'', ``did'', ``does'', ``is'', ``was''.
\end{itemize}
Temporal or relation cues enable graph retrieval, while single-hop questions without temporal or relation cues are kept on semantic retrieval whenever possible. In the hybrid setting, the rule-based plan is refined by an LLM only when the rule confidence is below a fixed threshold of 0.75. The refiner can update \texttt{use\_graph}, graph seed count, hop depth, and fusion weights, but its outputs are clipped to narrow intervals to keep the planner conservative and to avoid unconstrained benchmark-specific adaptation.

\subsection{Working Memory, Consolidation, and Context Reactivation}
Each participant maintains a FIFO working-memory queue with capacity $C_{\mathrm{w}}=20$. When the queue reaches capacity, AdaMem pops the oldest contiguous segment of $r=5$ messages and sends that segment to episodic-memory consolidation. This segment is selected purely by recency order; no query-aware filtering is used at write time. During retrieval, fact-based context reactivation uses a score-drop heuristic rather than a fixed context window: ranked fact matches are scanned in order, and supporting context is reactivated until the first adjacent score drop exceeds $\tau_{\mathrm{drop}}=0.1$.

\subsection{Topic Clustering and Persona Refresh}
We compute memory embeddings with \texttt{all-MiniLM-L6-v2} \citep{reimers2019sentence}. For both event/topic keys and attribute keys, AdaMem computes pairwise similarities, keeps only each key's most similar neighbor, and takes the connected components of this sparse similarity graph as merge clusters. Thus, clustering is sparse and threshold-free rather than based on a manually tuned similarity cutoff. Each event cluster is then passed to a topic-merge prompt that produces grouped topic memories, and the resulting grouped topics are further summarized into preference-oriented persona descriptors. Attribute clusters are merged into aspect-based persona summaries by a separate prompt.

\subsection{Graph Propagation Priors}
At indexing time, different edges receive fixed write-time strengths: message--topic \texttt{mentions} uses 0.75, message--fact \texttt{supports} uses 0.85, fact--event \texttt{supports} uses 0.80, message--attribute \texttt{supports} uses 0.80, same-speaker temporal edges use 0.90 forward and 0.45 backward, global temporal edges use 0.70 forward and 0.35 backward, and \texttt{speaker\_related} uses 0.65. At retrieval time, graph propagation uses the priors in Table~\ref{tab:app_edge_weights}. The planner can modestly increase the temporal prior for temporal questions and the speaker-related prior for attribute-heavy questions. Multi-hop propagation also uses a fixed hop-decay factor $\lambda=0.85$.

\begin{table}[t]
\centering
\caption{Default retrieval-time edge priors.}
\label{tab:app_edge_weights}
\begin{tabular}{lc}
\toprule
\textbf{Edge type} & \textbf{Weight} \\
\midrule
\texttt{mentions} & 0.75 \\
\texttt{supports} & 0.90 \\
\texttt{same\_topic} & 0.55 \\
\texttt{temporal\_next} & 0.70 \\
\texttt{speaker\_related} & 0.60 \\
\bottomrule
\end{tabular}
\end{table}

\subsection{Fusion Score Components and Weights}
The four fusion terms in Eq.~(3) are implemented as simple rank- and heuristic-based priors rather than learned scorers:
\begin{equation}
\begin{split}
s_{\mathrm{base}}(m,q) &= \frac{1}{1+\mathrm{rank}_{\mathrm{base}}(m)}, \\
s_{\mathrm{graph}}(m,q) &= \frac{1}{1+\mathrm{rank}_{\mathrm{graph}}(m)}.
\end{split}
\end{equation}
The recency term decreases linearly over the merged candidate list, and the factual term assigns a high bonus to messages already supported by baseline fact retrieval and a small residual bonus otherwise. For the benchmark configuration used in the released evaluation script, the default fusion prior is shown in Table~\ref{tab:app_fusion}. These values are fixed before evaluation rather than fit from supervision. The optional route refiner may adjust them per question, but only within clipped ranges: $\alpha \in [0.90,1.00]$, $\beta \in [0,0.08]$, $\gamma \in [0,0.03]$, and $\delta \in [0,0.03]$.

\begin{table}[t]
\centering
\caption{Default fusion prior in the benchmark configuration.}
\label{tab:app_fusion}
\begin{tabular}{lc}
\toprule
\textbf{Weight} & \textbf{Value} \\
\midrule
$\alpha$ semantic rank & 0.7 \\
$\beta$ graph rank & 0.1 \\
$\gamma$ recency & 0.1 \\
$\delta$ fact confidence & 0.1 \\
\bottomrule
\end{tabular}
\end{table}

\subsection{Representative Benchmark Defaults}
Table~\ref{tab:app_defaults} summarizes the main released retrieval defaults that are hidden behind the implementation in the public code. The number of Research Agent iterations is configured separately as an experimental variable and is therefore not treated as an immutable retrieval default here.

\begin{table}[t]
\centering
\caption{Representative released benchmark defaults.}
\label{tab:app_defaults}
\begin{tabular}{lc}
\toprule
\textbf{Setting} & \textbf{Value} \\
\midrule
Working-memory capacity $C_{\mathrm{w}}$ & 20 \\
Consolidation segment length $r$ & 5 \\
Baseline retrieval top-$K$ & 10 \\
Fact reactivation threshold $\tau_{\mathrm{drop}}$ & 0.1 \\
Base graph hop depth & 1 \\
Base graph seed count & 2 \\
\bottomrule
\end{tabular}
\end{table}

\section{Graph Construction Details}
\label{app:graph}

This appendix section clarifies how AdaMem builds and retrieves from graph memory. Importantly, the graph is not produced by a standalone LLM graph-extraction prompt. Instead, each utterance is first converted into a normalized record containing topic, facts, attributes, summary, and auxiliary fields; graph indexing then applies deterministic node typing and edge-construction rules on top of this record.

\subsection{Node and Edge Rules}
For each processed utterance, AdaMem creates a \texttt{message} node storing the raw text, speaker, timestamp, and turn index. The normalized fields create additional typed nodes: \texttt{topic} nodes for message topics, \texttt{fact} nodes for extracted factual snippets, \texttt{attribute} nodes for user- or counterpart-related attributes, and \texttt{event} nodes for event abstractions linked from facts or persona snapshots. We then add a small set of typed edges:
\begin{itemize}
    \item \texttt{mentions}: connects a message to its topic node;
    \item \texttt{supports}: connects a message to extracted fact or attribute nodes and connects fact nodes to their associated event nodes;
    \item \texttt{same\_topic}: connects the current message to a recent message from the same speaker when both mention the same topic, capturing local topical continuity;
    \item \texttt{temporal\_next}: connects each message to the immediately preceding message in the dialogue timeline and, separately, to the preceding message from the same speaker, providing lightweight temporal structure;
    \item \texttt{speaker\_related}: connects adjacent messages from the same speaker to preserve speaker-specific continuity.
\end{itemize}
Persona snapshots are also indexed as fact or attribute nodes so that long-term user descriptors remain reachable from graph retrieval.

\subsection{Graph Retrieval}
Given a question, AdaMem first selects semantic seed nodes and then performs bounded multi-hop expansion over typed edges. Expansion can be filtered by the inferred target referent, so that user-focused questions remain centered on user-grounded evidence while ambiguous questions can access both bundles. In the current implementation, propagation uses lightweight type-specific edge weights rather than learned graph reasoning; therefore, the graph should be understood as a structured retrieval scaffold rather than a fully learned knowledge-graph reasoner.

\section{Case Studies}
\label{app:case}

We analyze one representative success case in Figure~\ref{case1} and one representative failure case in Figure~\ref{case2} to better understand where AdaMem gains its advantage and where important challenges remain. 
\begin{figure}[htbp]
\centering

\setlength{\fboxsep}{6pt}
\fbox{
\begin{minipage}{0.95\linewidth}

\textbf{Question:} What activity did Caroline used to do with her dad?

\vspace{0.5em}
\textbf{Conversation Excerpt:}

\begin{ttfamily}
\footnotesize

Melanie: Oliver's hilarious! He hid his bone in my slipper once!
Cute, right? Almost as silly as when I got to feed a horse a carrot.

Caroline: That's so funny! I used to go horseback riding with my dad
when I was a kid. We'd go through the fields, feeling the wind.
It was so special. I've always had a love for horses!

Melanie: Wow, that sounds great – I agree, they're awesome.
Here's a photo of my horse painting I did recently.

\end{ttfamily}

\vspace{0.5em}

\textbf{Model Answers:}

\begin{itemize}

\item \textbf{Mem0:} No information available \xmark
\item \textbf{AdaMem:} horseback riding \cmark
\end{itemize}

\textbf{Reference Answer:} Horseback riding
\end{minipage}
}
\caption{\textbf{Case One (Success).}}
\label{case1}
\end{figure}

\paragraph{Why AdaMem succeeds.}
This case is deceptively simple: although the answer span appears explicitly in the dialogue, the question is not a pure surface-form lookup. The system must first identify that the target speaker is \textit{Caroline} rather than Melanie, then isolate the event embedded inside a broader horse-related exchange, and finally map the abstract query phrase  "what activity" to the concrete experience "go horseback riding with my dad." Mem0 fails because flat semantic storage is easily distracted by nearby but less relevant horse-related details (e.g., Melanie feeding a horse or showing a horse painting), and it lacks an explicit mechanism to prioritize the participant-specific evidence most aligned with the question. By contrast, AdaMem benefits from its participant-aware memory organization and normalized message writing: the utterance can be stored as a structured event/fact associated with Caroline, her childhood, and the father-related activity. During inference, target resolution narrows retrieval to Caroline's memory bundle, while topic-to-message recovery and relation-aware evidence aggregation restore the exact supporting utterance instead of only a vague semantic neighbor. The case therefore illustrates the central advantage of AdaMem: it succeeds not merely by retrieving similar text, but by preserving who said what, under which personal context, and how the event should be reconstructed for answering.

\begin{figure}[htbp]
\centering

\setlength{\fboxsep}{6pt}
\fbox{
\begin{minipage}{0.95\linewidth}

\textbf{Question:} When did Melanie read the book "nothing is impossible"?

\vspace{0.5em}
\textbf{Conversation Excerpt:}

\begin{ttfamily}
\footnotesize

Melanie: Wow, Caroline! You're so inspiring for wanting to help others
with their mental health. What's pushing you to keep going forward
with it?

Caroline: I struggled with mental health, and the support I got was
really helpful. It made me realize how important it is for others
to have a support system. So, I started looking into counseling
and mental health career options so I could help other people
on their own journeys like I was helped.

Melanie: Caroline, so glad you got the support! Your experience
really brought you to where you need to be. You're gonna make a
huge difference! This book I read last year reminds me to always
pursue my dreams, just like you are doing!

Caroline: Thanks so much, Mel! Seeing this pic just made me
appreciate my love of reading even more. Books guide me,
motivate me and help me discover who I am. They're a huge
part of my journey, and this one's reminding me to keep
going and never give up!

\end{ttfamily}

\vspace{0.5em}

\textbf{Model Answers:}

\begin{itemize}

\item \textbf{Mem0:} No evidence in memories \xmark
\item \textbf{AdaMem:} No evidence available \xmark

\end{itemize}

\textbf{Reference Answer:} 2022

\end{minipage}
}
\caption{\textbf{Case Two (Failure).}}
\label{case2}
\end{figure}

\paragraph{Why AdaMem still fails.}
The failure case exposes a harder form of long-horizon reasoning that is only partially addressed by our current design. To answer correctly, the model must solve two implicit alignment problems: it must link the title mention ``\textit{nothing is impossible}'' in the question to Melanie's deictic expression ``this book'' in the dialogue, and it must further normalize the relative temporal phrase ``last year'' into the absolute answer ``2022.'' If either link is missing, retrieval returns incomplete evidence and the final answer becomes ungrounded. AdaMem improves temporal reasoning substantially in the aggregate, but its current pipeline still relies mainly on retrieval planning, graph expansion, and lightweight temporal salience signals rather than explicit canonicalization of relative time expressions at write time. As a result, the memory may preserve a coarse fact such as Melanie having read an inspiring book ``last year,'' yet fail to store a stable symbolic binding between the book title and the absolute year required by the question. Once that canonical link is absent from memory, the Research Agent cannot recover it through additional search alone, and the Working Agent correctly defaults to abstention. This example suggests that the next bottleneck for AdaMem is not broad evidence coverage, but finer-grained temporal grounding and implicit entity linking, which is also consistent with the limitations discussed in the Limitations section.

\section{Prompt Templates}
\label{app:prompt}

We provide the most representative prompt templates used in AdaMem.

\subsection{Message Understanding Prompt}
\paragraph{Role.}
This prompt is the entry point of the Memory Agent. It converts each incoming utterance into a normalized memory record with topic, attitude, reason, facts, attributes, and a short summary, which is then written into working memory and used by downstream memory update modules.
\paragraph{Content.}
\begin{PromptVerbatim}
Perform topic tagging on this message from user.

Requirements:
1. Identify one primary topic/event in the message.
2. Infer the author's attitude toward the event.
3. Infer the reason behind the attitude.
4. Extract facts or events revealed by the message.
5. Extract user attributes revealed by the message.
6. Produce a one-sentence summary and a brief rationale.

Return JSON:
{
  "text": "original message",
  "tags": {
    "topic": ["..."],
    "attitude": ["Positive|Negative|Mixed"],
    "reason": ["..."],
    "facts": ["..."],
    "attributes": ["..."]
  },
  "summary": "...",
  "rationale": "..."
}

Message:
"{message}"
\end{PromptVerbatim}

\subsection{Episodic Memory Router Prompts}
\paragraph{Role.}
After message understanding, AdaMem uses three router prompts to update long-term episodic memory for \textit{events/topics}, \textit{facts}, and \textit{attributes}. These prompts determine whether a new memory item should be added, used to update an existing record, or ignored as redundant.
\paragraph{Content.}
\begin{PromptVerbatim}
You are a user profile updater that maintains a dictionary of existing
memory items. For each new message-derived item, analyze whether to update
the profile and respond in JSON format.

Input:
- Profile: existing topics / facts / attributes
- Message: a new structured item extracted from the latest utterance

Processing:
- Compare the new item with existing entries semantically.
- Choose one action:
  * UPDATE: a contradictory or revised entry already exists
  * ADD: the item is new
  * IGNORE: an equivalent entry already exists
- If there is no matched target, return "None".

Return JSON:
{
  "Action": "[UPDATE|ADD|IGNORE]",
  "Target": "matched entry or 'None'"
}
\end{PromptVerbatim}

\subsection{Topic Merge Prompt}
\paragraph{Role.}
This prompt groups multiple fine-grained event topics into higher-level topic-centric memories. It is important for compressing long interaction histories into reusable themes without discarding event semantics.
\paragraph{Content.}
\begin{PromptVerbatim}
Given a group of topics extracted from users' messages, analyze whether
these topics can be logically grouped under common themes.

Rules:
1. Only merge topics that talk about the same underlying event/theme.
2. Preserve the original meaning and context of each topic.
3. Extract as many common details as possible when naming the merged topic.
4. Do not reveal the user's attitude in the merged topic name.
5. Keep distinct concepts separate.

Input:
{
  "topic1": "summary sentence 1",
  "topic2": "summary sentence 2",
  ...
}

Return JSON:
{
  "Grouped Topics": {
    "NewTopicName1": ["original_topic1", "original_topic2"],
    "NewTopicName2": ["original_topic3"]
  },
  "Grouping Rationale": "Explanation of the grouping"
}
\end{PromptVerbatim}

\subsection{Route Refinement Prompt}
\paragraph{Role.}
This prompt is used by the optional route refiner after the rule-based planner proposes an initial retrieval plan. Its purpose is not to freely redesign the retrieval pipeline, but to make a conservative adjustment to graph usage, hop depth, seed count, and fusion weights while avoiding unnecessary graph expansion for simple questions.
\paragraph{Content.}
\begin{PromptVerbatim}
Refine a memory retrieval route plan for precise QA.
Prefer exact extraction for single-hop/temporal questions and avoid
unnecessary graph expansion.

Question:
{question}

Current plan JSON:
{current_plan}

Return only JSON with keys:
{
  "use_graph": true/false,
  "use_baseline": true/false,
  "graph_topn": integer,
  "hop_k": integer,
  "fusion_alpha": number,
  "fusion_beta": number,
  "fusion_gamma": number,
  "fusion_delta": number,
  "confidence": number
}
\end{PromptVerbatim}

\subsection{Research Integration and Reflection Prompts}
\paragraph{Role.}
These prompts implement the core loop of the Research Agent. The integration prompt merges newly retrieved evidence into a consolidated working summary, while the reflection prompts judge whether the current summary is sufficient and, if not, generate follow-up retrieval requests.
\paragraph{Content.}
\begin{PromptVerbatim}
[Integrate Prompt]
You are the IntegrateAgent. Merge newly retrieved evidence with the current
working notes to produce a consolidated factual summary relevant to the
QUESTION.

QUESTION:
{question}

EVIDENCE:
{evidence}

CURRENT RESULT:
{current_result}

Instructions:
1. Keep useful, on-topic information from CURRENT RESULT.
2. Add new, relevant, well-supported facts from EVIDENCE.
3. Remove off-topic content.
4. Prefer concrete details such as entities, dates, numbers, and events.
5. Resolve contradictions by preferring more specific or more recent evidence.
6. Use timestamps when answering temporal questions.

Return JSON:
{
  "content": "merged factual summary",
  "sources": ["source-1", "source-2", ...]
}
\end{PromptVerbatim}

\begin{PromptVerbatim}
[Info Check Prompt]
You are the InfoCheckAgent. Judge whether the collected information is
sufficient to answer the QUESTION.

QUESTION:
{question}

RESULT:
{result}

Return JSON:
{
  "enough": true | false
}

[Follow-up Request Prompt]
You are the FollowUpRequestAgent. Generate focused follow-up retrieval
queries for the missing information.

QUESTION:
{question}

RESULT:
{result}

Instructions:
1. Identify what is still missing.
2. Generate 1-3 targeted retrieval queries.
3. Mention concrete entities or events whenever possible.

Return JSON:
{
  "new_requests": ["query-1", "query-2", ...]
}
\end{PromptVerbatim}

\subsection{Working Agent Answer Prompt}
\paragraph{Role.}
This prompt is used by the Working Agent to convert the research summary into the final answer. It explicitly encourages concise, entity-centric responses and injects additional persona attributes or factual snippets when available.
\paragraph{Content.}
\begin{PromptVerbatim}
[System Prompt]
You are role-playing as {speaker_b} in a conversation with {speaker_a}.
Your task is to answer questions about {speaker_a} or {speaker_b} in an
extremely concise manner based on the provided research summary.
Any content referring to 'User' refers to {speaker_a}.
Answer as concisely as possible and try to deduce clear answers rather than
return vague statements.

[User Prompt]
<RESEARCH SUMMARY>
{research_summary}

{extra_context}

The question is: {question}

Please only provide the content of the answer.
For date questions, use a specific format such as "15 July 2023" whenever
possible. For duration questions, answer in years, months, or days.
Generate answers primarily composed of concrete entities.
\end{PromptVerbatim}

\end{document}